# A Comparative Study of Pretrained Language Models for Long Clinical Text

Yikuan Li [1], Ramsey M. Wehbe, MD [2,3], Faraz S. Ahmad, MD [1,2,3], Hanyin Wang [1], Yuan Luo, PhD [1] [*]

Emails:
yikuan.li@northwestern.edu
ramsey.wehbe@northwestern.edu
faraz.ahmad@northwestern.edu
hanyin.wang@northwestern.edu
yuan.luo@northwestern.edu

[1] Division of Health and Biomedical Informatics, Department of Preventive Medicine, Feinberg School of Medicine, Northwestern University, Chicago, Illinois, USA
[2] Division of Cardiology, Department of Medicine, Feinberg School of Medicine, Northwestern University, Chicago, Illinois, USA
[3] Bluhm Cardiovascular Institute Center for Artificial Intelligence, Northwestern Medicine, Chicago, Illinois, USA

[*] Corresponding author

Word Count = 3,320

# Abstract

**Objective:** Clinical knowledge enriched transformer models (e.g., ClinicalBERT) have state-of-the-art results on clinical NLP (natural language processing) tasks. One of the core limitations of these transformer models is the substantial memory consumption due to their full self-attention mechanism, which leads to the performance degradation in long clinical texts. To overcome this, we propose to leverage long-sequence transformer models (e.g., Longformer and BigBird), which extend the maximum input sequence length from 512 to 4096, to enhance the ability to model long-term dependencies in long clinical texts.

**Materials and Methods:** Inspired by the success of long sequence transformer models and the fact that clinical notes are mostly long, we introduce two domain enriched language models, Clinical-Longformer and Clinical-BigBird, which are pre-trained on a large-scale clinical corpus. We evaluate both language models using 10 baseline tasks including named entity recognition, question answering, natural language inference, and document classification tasks.

**Results:** The results demonstrate that Clinical-Longformer and Clinical-BigBird consistently and significantly outperform ClinicalBERT and other short-sequence transformers in all 10 downstream tasks and achieve new state-of-the-art results.

**Discussion:** Our pre-trained language models provide the bedrock for clinical NLP using long texts. We have made our source code available at https://github.com/luoyuanlab/Clinical-Longformer, and the pre-trained models available for public download at: https://huggingface.co/yikuan8/Clinical-Longformer.

**Conclusion:** This study demonstrates that clinical knowledge enriched long-sequence transformers are able to learn long-term dependencies in long clinical text. Our methods can also inspire the development of other domain-enriched long-sequence transformers.

# Introduction

Transformer-based models have been wildly successful in setting state-of-the-art benchmarks on a broad range of natural language processing (NLP) tasks, including question answering, document classification, machine translation, text summarization, and others [1-3]. These successes have been replicated in the clinical and biomedical domain via pre-training language models using large-scale clinical or biomedical corpora, then fine-tuning on a variety of clinical or biomedical downstream tasks, including computational phenotyping [4], automatical ICD coding [5], knowledge graph completion [6] and clinical question answering [7].

The self-attention mechanism [8] is one of the most critical components that lead to the success of transformer-based models, which allows each token in the input sequence to independently interact with every other token in the sequence in parallel. However, the memory consumption of the self-attention mechanism grows quadratically with sequence length, resulting in impracticable training time, and easily reaching the memory limits of modern GPUs. Consequently, transformer-based models that leverage a complete self-attention mechanism, such as BERT and RoBERTa, typically have an input sequence length limit of 512 tokens. To deal with this limit when modeling long texts using transformer-based models, the input sequence shall be either truncated to the first 512 tokens or processed via a sliding window of 512 tokens with or without overlap. If the latter method is applied to a document-level classification task, an aggregation operation will be added to yield the final output from multiple snippets. Both methods ignore long-term dependencies spanning over 512 tokens and may achieve suboptimal results due to information loss. Additionally, this input token limitation of the self-attention mechanism could impact language model pre-training and then be amplified to downstream tasks. In clinical NLP, transformer-based modeling approaches have also encountered this limitation [9]. For example, the discharge summaries in MIMIC-III, which are often used to predict clinically meaningful events like hospital re-admission [10] or mortality [11], have 2,984 tokens (1,435 words) on average, far exceeding the 512 token limit of most full attention-based transformer models.

Recently, investigators have developed novel variants of transformers specifically for long sequences that reduce memory usage from quadratic to linear scale of the sequence length [12-14]. The core idea behind these models is to replace the full attention mechanism with a sparse attention mechanism, which is typically a blend of sliding windows and reduced global attention. These models are capable of processing up to 4,096 tokens and have empirically boosted performance on NLP tasks, including question answering as well as text summarization. However, to the best of our knowledge, long sequence transformers in the clinical and biomedical domain have not yet been systematically explored. The purpose of this manuscript is to examine the adaptability of these long sequence models to a series of clinical NLP tasks. We make the following contributions:

- We leverage large-scale clinical notes to pre-train two new language models, namely Clinical-Longformer and Clinical-BigBird.
- We demonstrate that both Clinical-Longformer and Clinical-BigBird improve the performance of a variety of downstream clinical NLP datasets, including question answering, named entity recognition, and document classification tasks.

## Background and Significance

*Clinical and Biomedical Transformers*

Transformer-based models, especially BERT [2], can be enriched with clinical and biomedical knowledge through pre-training on large-scale clinical and biomedical corpora. These domain-enriched models, for example, BioBERT [15] pre-trained on biomedical publications and ClinicalBERT [16] pre-trained on clinical narratives, set state-of-the-art benchmarks on downstream clinical and biomedical NLP tasks. Inspired by the success of these domain-enriched models, more pre-trained models were released to boost the performance of NLP models when applied to specific clinical scenarios. For example, Smit et al. [17] proposed CheXbert to annotate thoracic disease findings from radiology reports, which outperformed previous rule-based labelers with statistical significance. The model was pre-trained using a combination of human-annotated and machine-annotated radiology reports. He et al. [18] introduced DiseaseBERT, which infused disease knowledge to the BERT model by pre-training on a series of disease description passages that were constructed from Wikipedia and MeSH terms. DiseaseBERT achieved superior results on consumer health question answering tasks compared with BERT and ClinicalBERT. Michalopoulos et al. [19] proposed UmlsBERT, which used the Unified Medical Language System (UMLS) Metathesaurus to augment the domain knowledge learning ability of ClinicalBERT. Zhou et al. [20] developed CancerBERT to extract breast cancer-related concepts from clinical notes and pathology reports. Agrawal et al. [21] leveraged order contrastive pre-training on longitudinal data to tackle the difficulty when only a small proportion of the clinical notes were annotated. However, all models mentioned above were built on the vanilla BERT architecture, which has a limitation of 512 tokens in the input sequence length. This limitation may result in the information loss of long-term dependencies in the training processes.

*Transformers for Long Sequences*

Various attention mechanisms have been proposed to handle the large memory consumption of the attention operations in the vanilla transformer architecture. Transformer-XL [22] segmented a long sequence into multiple small chunks and then learned long-term dependencies with a left-to-right segment-level recurrence mechanism. Transformer-XL learns 5.5 times longer dependencies than the vanilla transformer models but loses the advantage of bidirectional representation of BERT-like models. In another study, Reformer [23] applied two techniques to reduce the complexity of transformer architecture by replacing dot-product attention operation with locality-sensitive hashing and sharing the activation function among layers. Reformer was able to process longer sequences at a faster speed and be more memory efficient. However, this enhancement improves space, time, and memory efficiency, but not accuracy on specific tasks. Almost simultaneously, Longformer [13] and BigBird [14] were proposed to drastically alleviate the memory consumption of transformer models by replacing the pairwise full attention mechanisms with a combination of sliding window attention and global attention mechanisms. They are slightly different regarding implementation and configuration of the global and local attention mechanism, where BigBird introduces additional contrastive predictive coding to train global tokens [14]. Both models support input sequences up to 4.096 tokens long (8 times the input sequence limit of BERT) and significantly improve performance on long-text question answering and summarization tasks. However, the adaptability of these long sequence transformers to the clinical and biomedical fields, where document length mostly exceeds the limits of BERT-like models, has not been investigated.

## Materials and Methods

In this section, we first introduce the clinical dataset we use as the pre-training corpus, followed by the pre-training processes for Clinical-Longformer and Clinical-BigBird. Next, we enumerate the downstream tasks we use to compare our long sequence models with the short sequence models. We also provide the technical details of pre-training and fine-tuning for the purposes of reproducing our results. The entire pipeline can be found in **Figure** *1*.

**Figure 1**: The pipeline for pre-training and fine-tuning transformer-based language models.

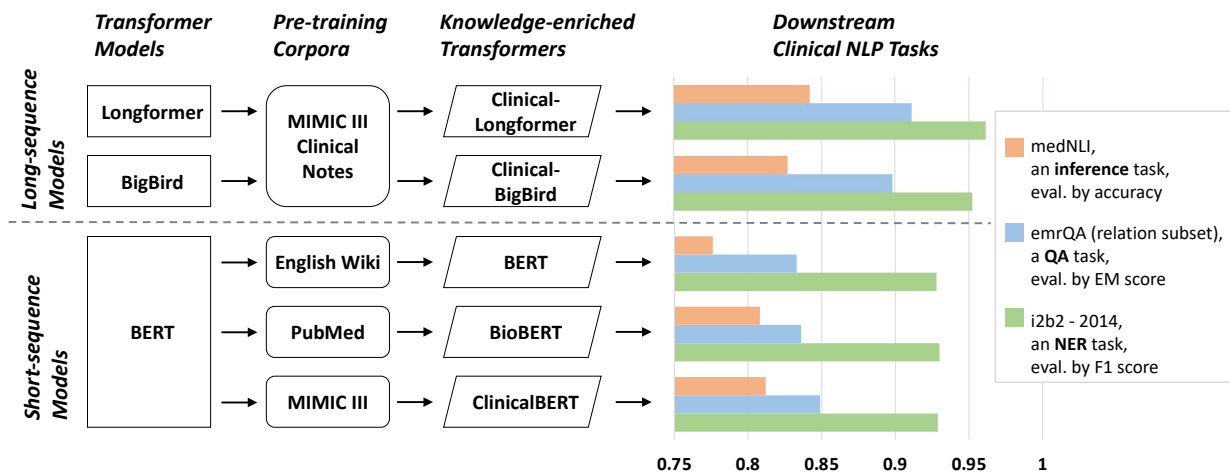

## *Datasets*

Similar to Huang et al. [10] and Alsentzer et al. [16], we leverage approximately 2 million clinical notes extracted from the MIMIC-III [24] dataset, which is the most extensive publicly available electronic health records (EHR) dataset that contains clinical narratives of over 40,000 patients admitted to the intensive care units. We only apply minimal pre-processing steps, including 1) to remove all de-identification placeholders from the clinical notes that were generated to protect the PHI (protected health information); 2) to remove all characters other than alphanumerics and punctuation marks; 3) to convert all alphabetical characters to lower cases, and 4) to strip extra white spaces. We believe that complicated pre-processing in the pre-training stage may not improve downstream performance but will sacrifice the generalizability of language models and significantly increase training time.

## *Pre-training*

Longformer [13] and BigBird [14] are the two best-performing transformer models that are designed for long input sequences. Both models extend the maximum input sequence length to 4,096 tokens, which is 8× the limit of conventional transformer-based models, by introducing localized sliding windows and global attention mechanisms to reduce the computational expenses of full self-attention mechanisms. The differences between the two models are how the global attention is realized and the selection of loss function in fine-tuning [13]. BigBird also contains some random localized attention operations. The reported performance difference between the two models on downstream tasks is minimal [14]. Therefore, we seek to pre-train both models and compare their performance on clinical NLP tasks. We refer readers to the original papers of Longformer [13] and BigBird [14] for more technical details.

We initialize Clinical-Longformer and Clinical-BigBird from the pre-trained weights of the base version of Longformer and the ITC (internal transformer construction) version of BigBird, respectively. Although the ETC (extended transformer construction) version of BigBird may have superior performance, HuggingFace (the largest community for sharing open-source pre-trained transformer models) only provides the implementation and the pre-trained checkpoints of the ITC version. The difference between ITC and ETC version is that in ITC version some existing tokens are made "global" and attend over the entire sequence, while ETC version introduces additionally "global" tokens such as CLS. Byte-level BPE (Byte-Pair-Encoding) [25] is applied to tokenize the clinical corpus. Both models are distributed in parallel to six 32GB Tesla V100 GPUs. FP16 precision is enabled to accelerate training. Batch size is 18 for Clinical-Longformer and 12 for Clinical-Bigbird, which are the upper limits under 6 32GB GPUs. We pre-train Clinical-Longformer for 200,000 steps and Clinical-BigBird for 300,000 steps, which ensures that each clinical note is seen equal times by the two models. The learning rates are 3e-5 for both models, the same as the learning rate used in the pretraining of Longformer. The entire pre-training process takes more than two weeks for each model.

To evaluate the performance of pre-training, we create a testing set that contains 1,000 documents that are also from MIMIC-III but have not been used as the pre-training corpora. Each document in the testing set is truncated to 512 tokens long. We randomly select 10% tokens from each document and replace them with a mask token. We compare our two pre-trained models with the short-sequence models in filling in the masked tokens using context. We report the perplexity score and top 5 accuracy in filling in the masked tokens of each model.

## *Downstream Tasks*

In this study, we fine-tune the pre-trained Clinical-Longformer and Clinical-BigBird on 10 clinical NLP datasets. These 10 NLP datasets broadly cover various NLP tasks, including extractive question answering, named entity recognition, natural language inference, and document classification. We rely on these NLP tasks to validate the performance improvement of long sequence models compared to their short sequence counterparts. The statistics and descriptions of all datasets can be found in Table 1.

### *Question Answering*

Question answering (QA) is a common NLP task that aims to automatically answer questions asked in natural language [26]. In the clinical context, QA systems answer clinicians' questions by understanding the clinical narratives extracted from electronic health record systems to support decision-making. emrQA [27] is the most frequently used benchmark dataset in clinical QA, which contains more than 400,000 question-answer pairs semi-automatically generated from past Informatics for Integrating Biology and the Bedside (i2b2) challenges. emrQA falls into the category of extractive question answering, aiming to identify answer spans from reference texts instead of generating new answers in a word-by-word fashion. Researchers have attempted to solve emrQA tasks by using word embedding models [28], conditional random fields (CRFs) [29] and transformer-based models [30], among which transformer-based models performed best. In our experiments, we investigate the performance of our pre-trained models using the three largest emrQA subsets: *Medication*, *Relation,* and *Heart Disease.* We evaluate QA performance with two commonly used metrics: exact match (EM) and F1-score. Exact match evaluates whether entire predicted spans match exactly with the ground-truth annotations. F1-score is a looser metric derived from token-level precision and recall, which measures the overlap between the predictions and the targets. We generate train-dev-test splits by following the instruction of Yue et al. [28]. The training set of *relation* and *medication* subsets are randomly under-sampled to reduce training time. Based on their experience, performance was not compromised by under-sampling. Of note, the emrQA dataset has some known issues, e.g., incomplete answers, it is template-based, and the annotation were generated semi-automatically[28]. We consider the usage of emrQA as a proof-of-concept experiment to compare the performance of the transformer-based model on the QA task.

Table 1: Description and statistics of downstream clinical NLP tasks

| Dataset | Task | Source | Sample Size | Avg. Seq. Length | Max Seq. Length |
|---|---|---|---|---|---|
| MedNLI | Inference | MIMIC | 14,049 | 39 | 409 |
| i2b2 2006 | NER | i2b2 | 66,034 | 867 | 3,986 |
| i2b2 2010 | NER | i2b2 | 43,947 | 1,459 | 6,052 |
| i2b2 2012 | NER | i2b2 | 13,108 | 794 | 2,900 |
| i2b2 2014 | NER | i2b2 | 83,466 | 5,134 | 14,370 |
| emrQA-Relation | QA | i2b2 | 255,908 | 1,880 | 6,109 |
| emrQA-Medication | QA | i2b2 | 141,243 | 1,460 | 6,050 |
| emrQA-HeartDisease | QA | i2b2 | 30,731 | 5,293 | 14,060 |
| openI | Multilabel Classif. | IndianaU | 3,684 | 70 | 294 |
| MIMIC-CXR | Multilabel Classif. | MIMIC-CXR | 222,713 | 119 | 874 |
| MIMIC-AKI | Binary Classif. | MIMIC | 16,536 | 1,463 | 20,857 |

*Named Entity Recognition*

Named entity recognition is a token-level classification task that seeks to identify the named entities and classify them in predefined categories. This genre of NLP tasks has broad applications in the clinical and biomedical domains. e.g., de-identification of PHI and medical concept extraction from clinical notes. Prior studies have shown that transformer-based models [15] significantly outperformed the models built on pre-trained static word embeddings [31] or LSTM networks [32]. We compare our pre-trained models using four data challenges: 1) i2b2 2006 [33] to de-identify PHI from medical discharge notes; 2) i2b2 2010 [34] to extract and annotate medical concepts from patient reports; 3) i2b2 2012 [35] to identify both clinical concepts and events relevant to the patient's clinical timeline from discharge summaries, and 4) i2b2 2014 [36] to identify PHI information from longitudinal clinical narratives. We follow the processing steps of Alsentzer et al. [16], which converts the raw data from all four tasks to the IOB (inside–outside–beginning) tagging format proposed by Ramshaw et al. [37], and then create train-dev-test splits. We evaluate the model performance with F1 score similarly to QA tasks.

*Document Classification*

Document classification is one of the most common NLP tasks, where a sentence or document is assigned to one or more classes or categories. In the clinical domain, document classification can be used to identify the onset of a particular disease process or predict patient prognosis using entire clinical notes. We use the following three document classification datasets to evaluate the pre-trained models from different perspectives.

*MIMIC-AKI* [38, 39] MIMIC-AKI is a binary classification task, where we aim to predict the possibility of AKI (acute kidney injury) for critically ill patients using the clinical notes within the first 24 hours following intensive care unit (ICU) admission. We follow Li et al. [38] to extract the cohort from MIMIC-III. We evaluate the model performance using AUC and F1 score.

*OpenI* [40] OpenI is a publicly available chest x-ray (CXR) dataset collected by Indiana University. The dataset provides around 4,000 radiology reports and their associated human annotated Medical Subject Headings (MeSH) terms. In our experiments, the task is to detect the presence of the annotated thoracic findings from CXR reports, which is considered a multi-label classification task. Given the small sample size, we will only use OpenI as the testing set. The pre-trained language models are fine-tuned using MIMIC-CXR [41], another publicly available chest x-ray dataset that contains more than 200,000 CXR reports. Unlike openI, the ground-truth labels for MIMIC-CXR were automatically generated using NLP approaches. The overlapping findings between the two CXR data sources are:: *Cardiomegaly*, *Edema*, *Consolidation*, *Pneumonia*, *Atelectasis*, *Pneumothorax* and *Pleural Effusion*. We report the sample number weighted average of the area under the receiver operating characteristic curve (AUC) as proposed and used in Li et al. [42]and Wang et al.[43].

*MedNLI* [44] Natural language inference (NLI) is the task of determining the relationship between sentence pairs. MedNLI is a collection of sentence pairs extracted from the MIMIC-III [24] and annotated by two board-certified radiologists. The relationship between the premise sentence and the hypothesis sentence could be entailment, contradiction, or neutral. Transformer-based models process NLI tasks also as document classification by merging the sentence pair and placing a delimiter token right after the end of the first sentence. We follow the original splits as Romanov et al. [44] and use accuracy to evaluate the performance.

## *Baseline Models and Comparisons*

Both Clinical-Longformer and Clinical-BigBird are compared with the short sequence models, including BERT, ClinicalBERT, RoBERTa, and BioBERT. We do not include the static word embedding models, e.g., Word2Vec and FastText, in the comparisons, because those models yield less competitive performance compared to the transformer-based models and cannot easily handle token-level classification tasks. The BERT [2] model is the first-of-its-kind transformer architecture that achieved state-of-the-art results on eleven NLP tasks. Both masked language modeling and next sentence prediction were used to learn contextualized word representation from BooksCorpus and English Wikipedia in the pre-training stage. BioBERT [15] is the first biomedical domain-specific BERT variant pre-trained from PubMed abstracts and PMC full-text articles. The weights of BioBERT were initialized from BERT. BioBERT yielded optimal performance in biomedical QA, NER, and relation extraction tasks. ClinicalBERT [16], initialized from BioBERT, was further pre-trained using clinical notes also extracted from MIMIC-III. ClinicalBERT boosted the performance for MedNLI and four i2b2 NER tasks that are also included in our study. BioBERT and ClinicalBERT use the next sentence prediction and masked language modeling as pre-training strategies. RoBERTa [3] is an improved variant of BERT model, which is trained with larger corpus, bigger batch size and gets rid of the next sentence prediction objectives. Both Longformer and BigBird initialize their training weights from RoBERTa checkpoint. We also try hierarchical transformers [45] in the experiment of MIMIC-AKI. The hierarchical transformer model uses the BERT model to learn outputs from each small chunk of text. Then, the outputs of small chunks are fed into the recurrent neural network. Given that the hierarchical transformer model is not explicitly developed for clinical NLP, we load the weights of ClinicalBERT to initialize the BERT layers.

*Experimental setup*

For the token-level classification, including QA and NER, a classification head is added to the output of each token obtained from the transformer-based architecture. The sequences are split into chunks in the length of 4,096 for Clinical-Longformer and Clinical-BigBird, and 512 for all the other three baseline models. 1,024 strides are taken between the chunks of long-sequence models; 128 strides are taken between the chunks of short-sequence models.

For the document classification tasks, the predicted outcomes are derived from the [CLS] token added to the beginning of each sentence or document. The maximum sequence lengths of OpenI and MedNLI are less than 512 tokens. Therefore, no truncation or sliding window approaches are needed for these two datasets. In MIMIC-AKI, given that some clinical notes are extremely long, which may even exceed the length limits of all models, we first truncate each document to the first 4096 tokens, which meets the length limits of Clinical-Longformer and Clinical-BigBird. The predicted outcomes are directly derived from the [CLS] output when using both long-sequence models. When dealing with short-sequence models, the documents are further segmented to snippets of 512 tokens in order to accommodate for the length requirement of short sequence models. A pooling strategy, which was introduced by Huang et al. [10] to predict ICU readmission from discharge summaries, is applied to aggregate the probability outputs from short snippets. The probability of AKI onset for a patient with $n$ short snippets is computed by:

$$P_{AKI} = \frac{[\max_{i \in n} p_i]^n + \frac{n}{2} * \left[\frac{1}{n} \sum_i^n p_i\right]^n}{1 + \frac{n}{2}}$$

, where $p_i$ is the probability output of the $i^{th}$ snippet from the short-sequence model. Our preliminary experiments show that this pooling strategy slightly outperforms the maximum pooling method.

We conduct our experiments using four 32GB GPUs. We maximize the batch size for each experiment given the memory limits of GPUs to save training time. The batch size during training is 16 for Clinical-Longformer, 12 for Clinical-BigBird, and 64 for all other models. Batch sizes are doubled when evaluating or testing. Half precision is applied to both Clinical-Longformer and Clinical-BigBird. We try learning rates: {1e-5, 2e-5, and 5e-5} for the experiments of each model on each task. We fine-tune 6 epochs for each set-up. All experiments converge within 6 epochs. The best-performing model parameters are determined by the performance of the development split. The experiments are implemented with python 3.8.0, PyTorch 1.9.0 and Transformer 4.9.0. The versions and downloadable links for all models can be found in Table S1.

## Results and Discussion

The evaluation of pre-training can be found in Table 2. The results demonstrate that Clinical-Longformer and Clinical-Bigbird can learn more useful contextualized relationships from clinical notes in the pre-training when compared to other baseline models, which provides the foundation for performance improvement in downstream tasks. BERT, BioBERT, and RoBERTa which are not pre-trained using clinical notes, yield very poor perplexity scores and MLM accuracies. This confirms that pre-training using domain-specific corpus is essential for learning the domain-specific contextualized relationships. We also visualize an example in Supplementary Figure 1. When [Stroke] is replaced with a mask token, Clinical-Longformer can infer this word from [infarct], [hemorrhagic], [epilepticus], and [hemorrhage], that are more than 1,000 tokens away from the mask token. This example demonstrates that our models can learn long-term dependencies from clinical narratives.

Full results for QA, NER, and classification tasks are presented in Table 3, 4, and 5, respectively (for full results with variance measurements, please see Table S2, S3, and S4). In question answering tasks, both Clinical-Longformer and Clinical-BigBird outperform the short-sequence transformer models by around 2 percent across all three emrQA subsets when evaluated by F1 score. When considering the stricter EM metric, Clinical-Longformer, and Clinical-BigBird improve ~ 5 percent on the relations subset but yield similar results to ClinicalBERT in the other two subsets. In NER tasks, Clinical-Longformer consistently leads the short-sequence transformers by more than 2 percent in all 4 i2b2 datasets. Clinical-BigBird also performs better than ClinicalBERT and BioBERT in all NER experiments. In document classification tasks, our two long-sequence transformers achieve superior results compared to prior models on OpenI, MIMIC-AKI, and medNLI tasks.

We observe that Clinical-Longformer and Clinical-BigBird not only improve the performance of long sequences tasks but also short sequences. The maximum sequences of MedNLI and OpenI are smaller than 512 tokens, which can be entirely fed into the BERT-like models. However, the long sequence models still achieve better results. We attribute these improvements to the pre-training stages of Clinical-Longformer and Clinical-BigBird, where the language models can learn more long-term dependencies by extending the sequence length limit, thereby learning a richer contextualization of clinical concepts. We find more significant gains, however, when applying our two long-sequence models to the datasets with longer sequences. For example, the performance improvement is most dramatic on the i2b2 2014 dataset, which has the largest average sequence length in all 4 NER tasks (almost twice the other three subsets). Likewise, Clinical-Longformer more strongly improves the F1 score of the *heart disease* subset from emrQA. This suggests that Clinical-Longformer and Clinical-BigBird are also better at modeling long-term dependencies in downstream tasks. Moreover, in i2b2 2006 dataset, the models achieve superior results in identifying the PHI information from the clinical notes. However, all PHI placeholders are completely removed in the pre-processing step of pre-training. This confirms that the language models can be generalized to new tokens in downstream tasks that are unseen in pre-training stage. Finally, we also find that Clinical-Longformer yields slightly better results when compared to Clinical-BigBird, although the differences in most experiments are not statistically significant. Given that Clinical-BigBird also requires more fine-tuning time and memory costs, we recommend that future investigators apply our Clinical-Longformer checkpoint to their own tasks when resources are limited.

Our study has several limitations. Firstly, we only apply Longformer and BigBird to large-scale clinical corpus. In future iterations, we plan to release more pre-trained models for long sequences enriched with other biomedical corpora, e.g., PubMed and PMC publications. Also, we only pre-train the base cased version of Clinical-Longformer and Clinical-BigBird. We will publish the un-cased and large version at the next step. Secondly, another recent approach developed to address the memory problem of long sequences is simplifying or compressing the transformer architecture. In future work, we will compare this genre of transformers, e.g., TinyBERT [46], to our current long sequence models. Thirdly, we do not integrate Clinical-Longformer or Clinical-BigBird into an encoder-decoder framework due to the memory limits of our GPU cards. Therefore, experiments on generative tasks like text generation or document summarization are not included in this study. We intend to incorporate these tasks into future versions of these models as our computational capability evolves. Fourthly, the emrQA was annotated in a semiautomatic way without expert calibration. There are incorrect NER labels as mentioned in Yue et al. [28]. We will conduct the experiments on a large-scale human-annotated NER dataset should there be any availability. Finally, the vocabularies of Clinical-Longformer and Clinical-BigBird are inherited from the 5,000 sub-word units used in the RoBERTa [3] model that was developed for non-clinical corpus. We have no idea if other types of tokenizers or a clinical-adaptive vocabulary can boost the performance. Therefore, we will examine more combinations in future studies.

**Table 2**: The evaluation of transformer-based models after language modeling (LM) pretraining

| Pre-trained Models | Perplexity Score | MLM Accuracy |
|---|---|---|
| BERT | 52807.11 | 0.633 |
| BioBERT | 131176.11 | 0.001 |
| ClinicalBERT | 8.67 | 0.803 |
| RoBERTa | 1378.71 | 0.693 |
| Clinical-Longformer | <u>1.61</u> | **0.940** |
| Clinical-BigBird | **1.41** | <u>0.936</u> |

*The best scores are in bold, and the second-best scores are underlined.*

**Table 3**: The performance of transformer-based pre-trained models on question answering tasks.

| Pre-trained Models | emrQA-Medication | | emrQA-Relation | | emrQA-Heart Disease | |
|---|---|---|---|---|---|---|
| *metrics* | EM | F1 | EM | F1 | EM | F1 |
| BERT | 0.240 | 0.675 | 0.833 | 0.924 | 0.650 | 0.698 |
| BioBERT | 0.247 | 0.700 | 0.836 | 0.926 | 0.647 | 0.702 |
| ClinicalBERT | 0.297 | 0.698 | 0.849 | 0.929 | <u>0.666</u> | <u>0.711</u> |
| RoBERTa | 0.280 | 0.706 | 0.825 | 0.917 | 0.655 | 0.682 |
| Clinical-Longformer | **0.302** | **0.716** | **0.911** | **0.948** | **0.698** | **0.734** |
| Clinical-BigBird | <u>0.300</u> | <u>0.715</u> | <u>0.898</u> | <u>0.944</u> | 0.664 | <u>0.711</u> |

*The best scores are in bold, and the second-best scores are underlined.*

Table 4: The performance of transformer-based pre-trained models on NER tasks.

| Pre-trained Models | i2b2 2006 | i2b2 2010 | i2b2 2012 | i2b2 2014 |
|---|---|---|---|---|
| *metrics* | F1 | F1 | F1 | F1 |
| BERT | 0.939 | 0.835 | 0.759 | 0.928 |
| BioBERT | 0.948 | 0.865 | <u>0.789</u> | 0.930 |
| ClinicalBERT | 0.951 | 0.861 | 0.773 | 0.929 |
| RoBERTa | 0.956 | 0.851 | 0.767 | 0.930 |
| Clinical-Longformer | **0.974** | **0.887** | **0.800** | **0.961** |
| Clinical-BigBird | <u>0.967</u> | <u>0.872</u> | 0.787 | <u>0.952</u> |

*The best scores are in bold, and the second-best scores are underlined.*

Table 5: The performance of transformer-based models on document classification tasks.

| Pre-trained Models | OpenI | MIMIC-AKI | | medNLI |
|---|---|---|---|---|
| *metrics* | Accuracy | AUC | F1 | Accuracy |
| BERT | 0.952 | 0.545 | 0.296 | 0.776 |
| BioBERT | 0.954 | 0.717 | 0.372 | 0.808 |
| ClinicalBERT | 0.967 | 0.747 | 0.468 | 0.812 |
| RoBERTa | 0.963 | 0.708 | 0.358 | 0.808 |
| Hierarchical Transformer | - | 0.726 | 0.462 | - |
| Clinical-Longformer | **0.977** | **0.762** | **0.484** | **0.842** |
| Clinical-BigBird | <u>0.972</u> | <u>0.755</u> | <u>0.480</u> | <u>0.827</u> |

*The best scores are in bold, and the second-best scores are underlined.*

## Conclusion

In this study, we introduce Clinical-Longformer and Clinical-BigBird, two pre-trained language models designed specifically for long clinical text NLP tasks. We compare these two models with the BERT-variant short-sequence transformer-based models, e.g., ClinicalBERT, in named entity recognition, question answering, and document classification tasks. Results demonstrate that Clinical-Longformer and Clinical-BigBird achieve better results on both long sequence and short sequence benchmark datasets. Future studies will investigate the generalizability of our proposed models to clinical text generation and summarization tasks, and the comparison with other modeling approaches that are also developed to solve the memory consumption of long text.

## Disclosure

All Authors declare no Competing Financial or Non-Financial Interests.

## Author Contributions

## Funding

This work was supported by the National Institutes of Health [U01TR003528 and R01LM013337].

## Data Availability

The benchmark datasets are derived from multiple publicly available datasets, including MIMIC III from https://physionet.org/content/mimiciii/1.4/; MIMIC-CXR from https://physionet.org/content/mimic-cxr/2.0.0/; i2b2 from https://portal.dbmi.hms.harvard.edu/; openI from https://openi.nlm.nih.gov/; and MedNLI from https://physionet.org/content/mednli/1.0.0/. To officially gain access, the authors should apply and sign data user agreement with the data owner. We provide codes to pre-process and generate splits at: https://github.com/luoyuanlab/Clinical-Longformer.

23